\pdfoutput=1

\documentclass[11pt]{article}

\usepackage[]{./template/acl}

\usepackage{times}
\usepackage{latexsym}

\usepackage[T1]{fontenc}

\usepackage[utf8]{inputenc}

\usepackage{microtype}

\usepackage{ctable} 
\usepackage{subcaption}
\usepackage{amsmath,amsfonts,amssymb}
\usepackage{multicol}
\newsavebox{\algleft}
\newsavebox{\algright}
\usepackage{bm}
\usepackage[linesnumbered,ruled,vlined]{algorithm2e}
\usepackage{hyperref}       
\usepackage{url}            
\usepackage{booktabs}       
\usepackage{amsfonts}       
\usepackage{nicefrac}       
\usepackage{xcolor}         
\usepackage{multirow}

\graphicspath{{./figure/}}

\usepackage{stfloats}

\title{Domain-Aware Contrastive Knowledge Transfer for Multi-domain Imbalanced Data}

\author{Zixuan Ke\thanks{~~Work done as an intern at Amazon Alexa AI.} \\
  UIC Computer Science \\
  Chicago, IL \\
  \texttt{zke4@uic.edu} \\
  \And
  Mohammad Kachuee \\
  Amazon Alexa AI \\
  Seattle, WA \\
  \texttt{~~~~~~~~~~~~~~~~\{kachum,sungjinl\}@amazon.com} \\ 
  \And
  Sungjin Lee\\
  Amazon Alexa AI \\
  Seattle, WA \\
  }

\date{}

\begin{document}
\maketitle
\begin{abstract}
In many real-world machine learning applications, samples belong to a set of domains e.g., for product reviews each review belongs to a product category. 
In this paper, we study multi-domain imbalanced learning (MIL), the scenario that there is imbalance not only in classes but also in domains.
In the MIL setting, different domains exhibit different patterns and there is a varying degree of similarity and divergence among domains posing opportunities and challenges for transfer learning especially when faced with limited or insufficient training data.
We propose a novel domain-aware contrastive knowledge transfer method called DCMI to (1) identify the shared domain knowledge to encourage positive transfer among similar domains (in particular from head domains to tail domains); (2) isolate the domain-specific knowledge to minimize the negative transfer from dissimilar domains.
We evaluated the performance of DCMI on three different datasets showing significant improvements in different MIL scenarios.
\end{abstract}

\section{Introduction}
The majority of existing works in imbalanced learning focus on the \textit{class imbalance setting} where classes are presented in a long-tailed distribution: a subset of classes (head classes) have sufficient samples, while other uncommon or rare classes (tail classes) are underrepresented by limited samples. This setting is challenging because the model naturally focuses largely on the majority classes and there may be no sufficient data for tail classes to recover their underlying distribution \cite{liu2019large}.

Even though extensive work has been done on the class imbalance problem, the consideration of \textit{domains}\footnote{In this paper, the term \textit{domain} is used to refer to a segmentation of samples, and it should not be confused with the same term also used in the domain adaptation literature studying the distribution shift problem.} is often missed. In many real-world scenarios, data naturally belongs to a set of domains e.g., for an online store, a potential domain assignment for each customer review can be defined based on the corresponding store departments.

A simplistic solution is to ignore domain assignments and train a classifier for all domains, which we refer to as \textit{domain agnostic learning (D-AL)}. D-AL entirely ignores domains and assumes that the model can ``automatically'' discover the data distribution for domains and learn them equally well. The drawbacks of such an approach are obvious: if the training data is sourced from many domains, updating all parameters may lead the model to focus on the subsets of the data in proportion to their ease of access or frequency. Moreover, if the data from different domains are dissimilar, agnostic learning may cause undesirable convergence dynamics i.e., negative transfer. 
We, therefore, argue that in the \textit{multi-domain imbalanced learning (MIL)} scenarios, a learning algorithm should consider domain information and leverage them to achieve effective knowledge transfer.

The MIL is a challenging problem. First, different domains may have very different number of samples and show a long-tailed distribution. For example, an intelligent assistant (e.g. Amazon Alexa) may provide a wide variety of skills and different skills may vary largely in number of examples. It is possible that some internal developed skills (e.g. music or whether) have hundreds of thousands of samples while many third-party developed skills may have only less than 10 samples in the same dataset \cite{kachuee2021self}. Second, domains may exhibit different semantic similarities and disparities with each other. For instance, a feature may show positive correlation with a label for certain domains while it is negatively correlated for others.
Third, the data-provided domain annotation may not be completely accurate or sufficiently fine-grained. For example, a sentence \textit{``Due to software or hardware issues, my computer cannot open my favorite text book, One hundred Years of Solitude''} may belong to both \textit{computer} and \textit{books} domains while it may have only one domain assignment in the dataset.

Perhaps the most intuitive approach for MIL is \textit{multi-task learning (MTL)}, where separate heads are used for different domains. While MTL considers domains, we will show it performs poorly in our experiments due to the lack of knowledge transfer between the classifiers. We believe that the key to successful MIL is to not only enable but encourage positive transfer learning across domains.

In this paper, we propose \underline{D}omain-aware \underline{C}ontrastive Knowledge Transfer for \underline{M}uti-domain \underline{I}mbalance learning (DCMI).
DCMI introduces a novel \textit{domain-aware representation layer} based on domain embeddings which enables fine-grained and scalable representation sharing or separation.
Complementary to the data provided domain assignments, we use an auxiliary domain classification task to help determine the relevance of a \textit{sample} to each domain i.e., \textit{soft domain assignments}.
DCMI uses a novel contrastive knowledge transfer objective to move the representation from similar domains closer and representation from dissimilar domains further apart. We conduct extensive experiments on three different multi-domain imbalanced datasets to demonstrate the effectiveness of DCMI.



\section{Related Work}

The recent imbalance learning literature can be organized into the following categories:

\textbf{Data Resampling.} This is one of the most widely used practices to artificially balance the distribution. Two popular options are under-sampling \cite{buda2018systematic,more2016survey} and over-sampling \cite{buda2018systematic,sarafianos2018deep,shen2016relay}. Under-sampling removes data from the head (dominant classes) while over-sampling repeats the data from the tail (minority classes). These approaches can be problematic as discarding tends to remove important samples and duplicating tends to introduce bias or overfitting.

\textbf{Data Augmentation.}
Data augmentation has been used to enrich the tail classes.
A popular approach is to leverage the Mixup \cite{zhang2018mixup} technique to augment the minority classes. Remix \cite{chou2020remix} assigns the label in favor of minority classes to the mixup samples, \citet{liu2020deep} prepares a ``feature cloud'' for mixing up that has a larger distribution range for tail classes. 
\citet{kim2020m2m} adds noise to head classes to generate tail classes. 
\citet{chu2020feature} decomposes the feature spaces and generate tail classes samples by combining class-shared feature from head classes and class-specific features from tail classes. 
However, this is usually a non-trivial work to generate meaningful samples that can help tail classes.

\textbf{Loss Reweighting.}
The basic idea of reweighting is to allocate larger weight for loss terms corresponding to tail classes while less weight for head classes.
In class-sensitive cross-entropy loss \cite{japkowicz2002class}, the weight for each class is inversely proportional to the number of samples. 
\citet{ren2018learning} leverages a hold-out evaluation set to minimize the balanced loss.

\textbf{Regularization.} This approach adds an additional regularization term to improve the training for the tail samples. \citet{lin2017focal} adds a factor to the standard cross-entropy loss to put more focus on hard, misclassified samples (usually attributed to the minority classes). \citet{cao2019learning} proposed to regularize the minority classes strongly so that the generalization error of minority classes can be improved. While regularization is simple and effective, the soft penalty can be insufficient to make the model focus on the tail classes and a large penalty may negatively affect the learning itself.

\textbf{Parameter isolation.} It has been shown that decoupling the learning into representation learning and classifier learning can be quite effective.
BBN \citet{zhou2020bbn} proposed a two-branch approach where the representation learning branch is trained as there is no class imbalance (input random sampling data) while the classifier learning branch applies the reverse sampling technique. The two branches are then combined by a curriculum learning strategy. 
\citet{wang2021contrastive} further improves BBN by replacing the cross-entropy loss in representation learning branch into a prototypical supervised contrastive loss. This approach offers the opportunity to optimize each part separately but also make it hard to transfer knowledge from head to tail classes

\textbf{Domain Imbalanced Learning.} The above approaches mostly consider the class imbalance but ignore the imbalance across domains. \citet{cheng2020representation} proposed a doubly balancing technique for both class imbalance and cross-domain imbalance, which  only limited to two domains, without any explicit mechanism to encourage the positive transfer and avoid the negative transfer. 





\section{Problem Definition}
In this paper, we assume access to a set of samples $(\bm{x}_i,y_i,j)$ for $i=\{1 \dots N\}$, $ y_{i} \in \{1\dots C\}$, and $j \in \{1\dots M\}$. Here, $N$ is the number of samples,  $C$ is the number of classes, and  $M$ is number of domains, i.e., shared feature space and label set across domains. 
We assume the scenario where exists (a) \textit{class imbalance}: classes are not evenly distributed in each domain; (b) \textit{domain imbalance}: domains are not evenly distributed, i.e., some domains may have much more or less number of examples than other domains; and (c) \textit{domain divergence}: while some domains are naturally similar to others and thus positively correlated, some domains are naturally dissimilar to others and negatively correlated. Given these assumptions, in \textit{multi-domain imbalanced learning} (MIL) we seek a model to minimize the expected loss for all domains (i.e., macro average).

\section{Proposed Method}
\begin{figure}[t]
\centering
\includegraphics[width=\columnwidth]{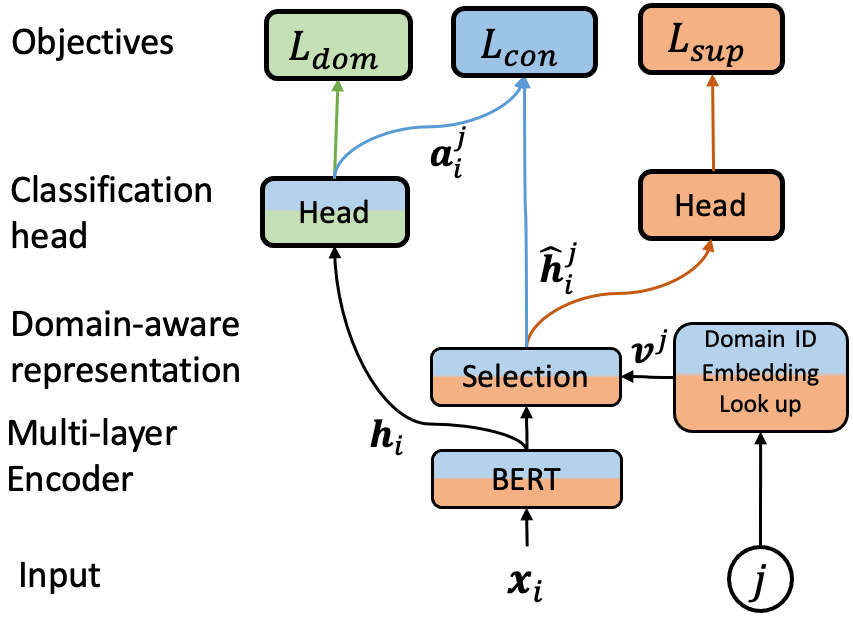}
\caption{An overview of the DCMI training process. 
$\bm{(i)}$ DCMI takes as input a sample $x^{(i)}$ from domain $j$.
$\bm{(ii)}$ The encoded feature vector $\bm{h}_i$ is computed using a shared body network (e.g., BERT).
$\bm{(iii)}$ The domain index is used to get the corresponding domain embedding used to compute the domain mask $\bm{m}_j$ and domain-aware representation $\bm{\hat{h}}^j_i$.
$\bm{(iv)}$ The supervised classification ($\mathcal{L}_{\text{sup}}$), contrastive ($\mathcal{L}_{\text{con}}$), and domain classification ($\mathcal{L}_{\text{dom}}$) loss terms are computed (see Section~\ref{sec:contrast}).
$\bm{(v)}$ The flow of gradients from each loss term is controlled such that each term is only used to optimize a subset of trainable parameters as indicated by green, blue, and orange colors in the drawing.}
\label{fig:overview}
\end{figure}

Fig.~\ref{fig:overview} presents an overview of the proposed method. 
In the MIL problem, it is crucial to identify the \textit{shared knowledge} that can be transferred across similar domains to improve the tail domain performance and the \textit{domain-specific knowledge} that needs to be handled carefully to avoid a negative transfer. 
To obtain domain-aware representations, we leverage domain embeddings to adaptively select the useful representation for each specific domain (Sec. \ref{sec:selection}). Additionally, regardless of the dataset provided domain assignment, in reality, a sample can belong to multiple domains to different degrees. To address this, we propose a \textit{domain classification} task to obtain the relevance of a sample to each domain and transfer the related domain knowledge using a \textit{contrastive} method (Sec. \ref{sec:contrast}).

\subsection{Domain-Aware Representation}
\label{sec:selection}




We suggest a domain-aware representation layer 
to adaptively select the appropriate representation (neurons) for each domain. For a domain $j$, the corresponding embedding $\bm{v}^j$ consists of differentiable parameters that can be learned in an end-to-end fashion. Based on this, the sigmoid function is used to find the corresponding domain mask $\bm{m}^j$:
\begin{equation}
\bm{m}^j = \sigma(\bm{v}^j/\tau) \;.
\end{equation}
Where $\tau$ is a temperature variable, linearly annealed from 1 to $\tau_{\min}$ (a small positive value).

To obtain the \textit{domain-aware representation}, we use element-wise multiplication of the output of the body network (i.e., BERT in this paper) $\bm{h}$ and the mask $\bm{m}^j$:
\begin{equation}
\bm{\hat{h}}^j_i = \bm{h}_i \odot \bm{m}^j \;.
\end{equation}
Note that the neurons in $\bm{m}^j$ may overlap with those in other domain masks to enable knowledge sharing. 

To make sure the $\bm{v}^j$ to have a wide range and its gradient to have a large magnitude, a gradient compensation technique is employed to the original gradient $\bm{g}$ \cite{Serra2018overcoming}. Specifically,
\begin{equation}
\label{eq:smax}
\bm{g}' = \frac{\tau[\text{cosh}(\bm{v}^j/\tau)+1]}{\tau_{\min}[\text{cosh}(\bm{v}^j)+1]} \odot \bm{g} \;.
\end{equation}

The embedding matrix is trained jointly with the supervised classification objective using a typical cross-entropy loss, denoted by $\mathcal{L}_{\text{sup}}$.

\subsection{Contrastive Knowledge Transfer}
\label{sec:contrast}

Even though we obtain the domain-aware representation using the suggested domain embedding, there are two limitations: (a) apart from supporting shared features, there is no explicit mechanism to actively encourage knowledge transfer; (b) the dataset provided domains are not necessarily accurate and fine-grained in the real world. Certain examples can be attributed to multiple domains with different degrees of relevance. For example, a review written on a product is usually considered in the general domain of that product (e.g., computers); however, semantically, it may involve discussion of other domains (e.g., the music playback quality of a laptop).

To address the above issues, we employ a domain classification task to estimate the relevance of each sample to different domains. 
We leverage these relevance/confidence scores as soft labels to conduct contrastive learning, allowing knowledge transfer from similar domains at the instance level.

\textbf{Domain Classification. } To estimate the relevance of different domains for a given sample, we leverage a sigmoid classification head with $M$ output neurons. For training, we employ binary cross-entropy (BCE) loss $\mathcal{L}_{\text{dom}}$ using the dataset provided domain assignments as labels.
Using the trained domain classifier, assuming it can generalize and capture domain similarities, we estimate the relevance of sample $i$ to each domain using its sigmoid output score for domain $j$, denoted by $\bm{a}^j_i$.

Note that the domain classification task is only an auxiliary task to be used in the contrastive learning objective explained next. Therefore, we block gradients from this objective to flow outside the domain classifier head.

\begin{figure}[t]
\centering
\includegraphics[width=\columnwidth]{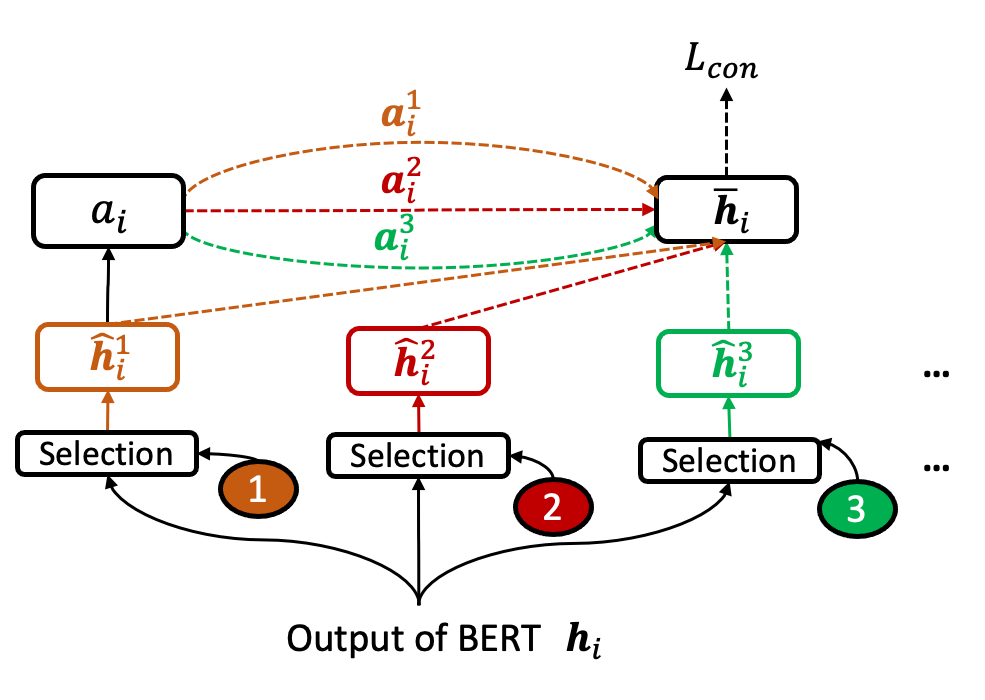}
\caption{An illustration of the contrastive learning objective:
$\bm{(i)}$ Domain-aware representations $\bm{\hat{h}}_i^j$ are computed for sample $i$ and all domains indexed by $j$.
$\bm{(ii)}$ Sigmoid outputs of the domain classifier head $\bm{a}^j_i$ are used to compute a weighted average of domain-aware representations resulting in an augmented view $\bm{\overline{h}}_i$.
$\bm{(iii)}$ A soft cross-entropy loss based on the augmented view and domain certainties is used as the contrastive objective function.}
\label{fig:contrast}
\end{figure}

\begin{figure}[t]
\centering
\includegraphics[width=\columnwidth]{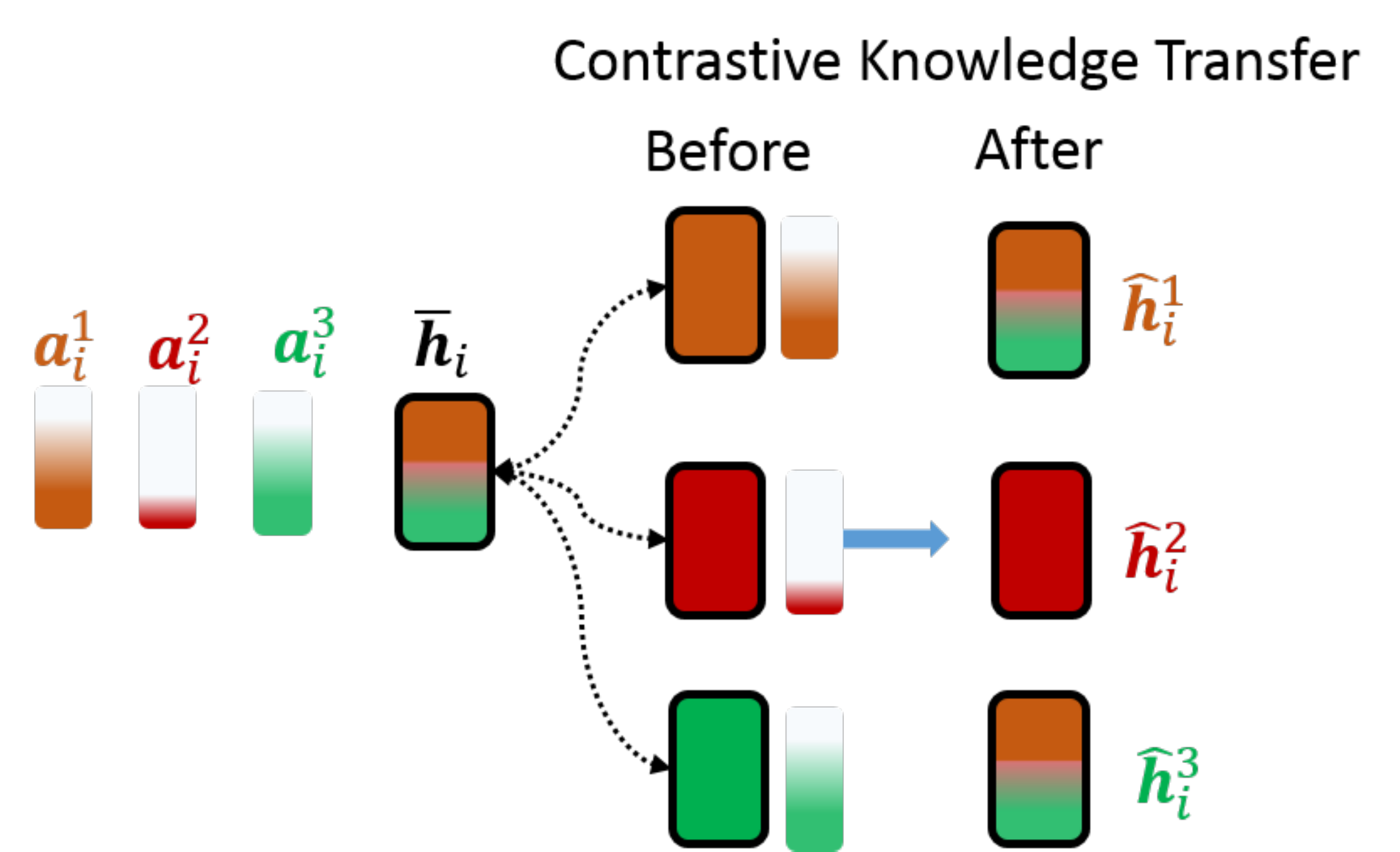}
\caption{A simple example to show the effectiveness of the contrastive knowledge transfer. Orange, red, and green bars show the degrees of relevance to domains 1,2, and 3, respectively. Here, the contrastive objective encourages similar domains (domain 1 and 3) to have similar representations, while the sample belonging to a dissimilar domain (domain 2) is pushed apart in the representation space.}
\label{fig:contrast_eg}
\end{figure}
\textbf{Contrastive Learning.} Fig.~\ref{fig:contrast} shows an illustration of the proposed contrastive objective. Here, for a certain sample, regardless of the dataset provided domain, we compute its domain-aware representations for all domains: $\bm{\hat{h}}^1_i$$\dots$$\bm{\hat{h}}^M_i$.
Then, we compute an augmented view of the sample by simply computing a weighted average of domain-aware representations and their normalized relevance:
\begin{equation}
\bm{\overline{h}}_i = \sum_{j=1}^M \frac{\bm{a}^j_i}{\sum_{j=1}^M \bm{a}^j_i} \bm{\hat{h}}^j \;.
\end{equation}

Based on this, we define the contrastive objective as:
\begin{multline}
    \label{eq:contrastive}
    \mathcal{L}_{\text{con}} = -\frac{1}{N}\sum_{i=1}^N\sum_{j=1}^M \bm{a}^j_i\log(\sigma(\bm{\overline{h}}_i\cdot \bm{\hat{h}}^j_i)) + \\
    (1-\bm{a}^j_i)\log(1-\sigma(\bm{\overline{h}}_i\cdot \bm{\hat{h}}^j_i))) \;,
\end{multline}
which is essentially a soft cross-entropy loss.
Intuitively, the contrastive objective of \eqref{eq:contrastive} encourages learning representations that capture the attribution of the augmented view to each domain. Through this objective, similar domains are represented with closer representations and dissimilar domains are moved further apart such that they are easily distinguishable from the augmented view. Note that $\mathcal{L}_{\text{con}}$ is different from the typical contrastive objectives usually used in the literature as it relies on soft domain assignments for the augmented view rather than distinguishing augmented and real data.

As an example, assume that the domain-aware representation $\bm{\hat{h}}^j_i$ is not a good representation for sample $i$ and lacks knowledge that is potentially transferable from other domains (indicating by a single color in their representation boxes), we can see how $ \mathcal{L}_{\text{con}}$ helps (see Fig. \ref{fig:contrast_eg}):
\begin{itemize}
    \item \textit{Sample $i$ semantically relevant to multiple domains} (domain 1 and domain 3). 
    In this case, $\bm{a}^1_i$ and $\bm{a}^3_i$ have a large value 
    while $\bm{a}^2_i$ has a smaller value. 
    Consequently, $\bm{\overline{h}}_i$ is mostly the average of $\bm{\hat{h}}^1_i$ and $\bm{\hat{h}}^3_i$ (half orange and half green). Here, updating based on $ \mathcal{L}_{\text{con}}$ moves $\bm{\hat{h}}^1_i$ and $\bm{\hat{h}}^3_i$ closer to $\bm{\overline{h}}$ 
    . In other words, the knowledge transfer is encouraged between the first and third representations for that sample. 
    \item \textit{Sample $i$ is not semantically relevant to a domain} (domain 2). Updating based on $ \mathcal{L}_{\text{con}}$, $\bm{\hat{h}}^2_i$ moves further from  $\bm{\overline{h}}_i$ to reflect the difference between them. Consequently, $\bm{\hat{h}}^2_i$ is discouraged from a negative knowledge transfer. 
    This is expected as $\bm{\hat{h}}^2_i$ is not relevant to sample $i$. 
\end{itemize}

\subsection{Implementation Details}
\textbf{Final Objective.}
The final joint training objective is a combination of the supervised classification, domain classification and sample level contrastive loss terms:
\begin{equation}
\label{eq.final}
\mathcal{L} =  \mathcal{L}_{\text{sup}} + \lambda_{\text{1}}\mathcal{L}_{\text{dom}} + \lambda_{\text{2}}\mathcal{L}_{\text{con}}, 
\end{equation}
where, $\lambda_1$ and $\lambda_2$ are hyperparameters to adjust the impact of each term.
Note that gradients computed from each objective update different parts of the network as shown in Fig. \ref{fig:overview} via different colors. For example, $\mathcal{L}_{\text{dom}}$ only updates the domain classifier head, and $\mathcal{L}_{\text{con}}$ updates all parameters except those in the supervised classification head.

\textbf{Architecture.}
A fully connected layer with softmax output is used as the classification head in the last layer of BERT. We use the embedding of \texttt{[CLS]} as the output of BERT. The training of BERT, follows that of \cite{DBLP:conf/naacl/XuLSY19}. We adopt $\text{BERT}_{\textbf{BASE}}$ (uncased).

\textbf{Hyperparameters.}
\label{sec:Hyperparameters}
Unless otherwise stated, the domain id embeddings have $768$ dimensions. We use $0.0025$ for $\tau_{\min}$ in Eq.~\ref{eq:smax}. A dropout layer with the rate of $0.5$ is placed between fully connected layers. To find the $\lambda_1$ and $\lambda_2$ hyperparameters in Eq.~\ref{eq.final}, we conducted a grid search in the $[0,5000]$ range using about 200 logarithmic increments. We provide the selected $\lambda_1$ and $\lambda_2$ for each dataset in Section~\ref{sec:baselines}.
For the contrastive objective, an $l_2$ normalization is applied before computing the contrastive loss.  The max length of the number of input tokens is set to 128. We use Adam optimizer and set the learning rate to $3\times10^{-5}$. For all experiments, we train for 5 epochs using a mini-batch size of 64.

\section{Experiments}

\subsection{Experimental Setup}
\subsubsection{Datasets}
\label{sec:datasets}
We conduct experiments using three datasets: \textit{Document Sentiment Classification (DSC)} \cite{ni2019justifying}, \textit{Aspect Sentiment Classification (ASC)} \cite{DBLP:conf/naacl/KeXL21} and \textit{Rumour and Fake News Detection (RFD)} \cite{DBLP:journals/corr/ZubiagaLP16,DBLP:conf/acl/Wang17}. These datasets have natural class and domain imbalance.
For all datasets, we use a random data split of 10\% for test, 10\% for validation, and the rest for training. To better evaluate the performance of each method in efficient knowledge transfer, we down-sample the training and validation sets of the DSC, ASC, and RFD with a factor of 1000, 10, and 10, respectively. We provide the exact domain and class statistics in the appendix. In addition to these datasets, we conduct additional experiments using an altered version of the ASC dataset with artificially dissimilar domains (Sec.~\ref{sec:asc_altered}).

\textbf{DSC.} 
For this dataset, the task is to classify each full product review into one of the two opinion classes (\textit{positive} and \textit{negative}). The training data provides the particular type of product being reviewed as domain information. We adopt the text classification formulation in \cite{DBLP:conf/naacl/DevlinCLT19}, where the \texttt{[CLS]} token is used to predict the opinion polarity.

To build the DSC dataset, we use 29 domains from the Amazon Review Datasets \cite{ni2019justifying} \footnote{\url{https://nijianmo.github.io/amazon/index.html}}, then binarize the ratings by converting 1-2 stars to negative and 4-5 stars to positive.

\textbf{ASC.} 
This dataset provides a classification of review sentences on their aspect-level sentiment (one of \textit{positive} and \textit{negative}).
For example, the sentence ``\textit{The picture is great but the sound is lousy}'' about a TV expresses a \textit{positive} opinion about the aspect ``picture'' and a \textit{negative} opinion about the aspect ``sound.'' 
We adopt the ASC implementation by \citet{DBLP:conf/naacl/XuLSY19}, where the aspect term and sentence are concatenated via \texttt{[SEP]} in BERT. The opinion is predicted using the \texttt{[CLS]} token.

The ASC dataset \cite{DBLP:conf/naacl/KeXL21} consists of 19 domains from 4 sources: $(a)$ \textit{HL5Domains} \cite{hu2004mining} with reviews of 5 products; $(b)$ \textit{Liu3Domains} \cite{liu2015automated} with reviews of 3 products; $(c)$ \textit{Ding9Domains} \cite{ding2008holistic} with reviews of 9 products; and $(d)$ \textit{SemEval14} with reviews of 2 products - SemEval 2014 Task 4 for laptop and restaurant. 

\textbf{RFD.} 
This dataset is compose of PHEME rumor detection \cite{DBLP:journals/corr/ZubiagaLP16} and LIAR fake news detection \cite{DBLP:conf/acl/Wang17} datasets. For rumor detection, the task is to identify whether a piece of given news is a rumor or not, while for the fake news detection, it is to identify fake or real news pieces. We follow \citet{DBLP:conf/naacl/DevlinCLT19} where the \texttt{[CLS]} token is used 
for the classification.

The RFD dataset consists of 6 domains from the PHEME dataset (5 domains) of rumor tweets \cite{DBLP:journals/corr/ZubiagaLP16}\footnote{\url{https://figshare.com/articles/dataset/PHEME_dataset_of_rumours_and_non-rumours/4010619}} and the fake news detection LIAR \cite{DBLP:conf/acl/Wang17} (1 domain).
Note that domains in PHEME defined by different news events (e.g. a specific shooting incident), while the domain in LIAR is defined by news genres (e.g. politics). We intentionally selected this dataset to evaluate the performance of different methods when domains are merely a segmentation of samples rather than following a consistent definition.

\subsubsection{Metrics}
For each experiment, we report Area Under the ROC Curve (AUC) as the performance measure. Two types of results are reported: \textit{macro} and \textit{micro}. Macro is computed by macro averaging results computed for individual domains. Micro is computed from averaging the performance of all test samples regardless of their domain assignments. Note that there is an imbalance in the frequency of class labels (positive and negative in ASC, DSC; fake and real in RFD) in addition to the imbalance in the domains for each dataset. To ensure the statistical significance of the results, each experiment is repeated 5 times using random seed and random initialization, reporting the mean and standard deviation of each result.

\subsubsection{Comparison Baselines}
\label{sec:baselines}
As the main focus of this study is the domain imbalance, to address class imbalance existing in our benchmarks, we adopt the existing DRS method \cite{cao2019learning} for all experiments.
In our comparisons, we use multi-task learning (MTL) and domain-agnostic learning (D-AL) as intuitive and straightforward baselines. 
Additionally, since little work has been done in MIL, we adapt the recent class imbalance systems to MIL by re-sampling or re-weighting based on the domain statistics. For each case, we follow similar architectures as DCMI to ensure fair comparisons. The compared methods cover various approaches including: loss re-weighting (D-DRW \cite{cao2019learning}), regularization (D-Focal \cite{lin2017focal}),  re-sampling (D-DRS \cite{cao2019learning}), parameter isolation (D-BBN \cite{zhou2020bbn} and D-HybridSC \cite{wang2021contrastive}), and mixture-of-experts (D-MDFEND \cite{nan2021mdfend}).
Note that the prefix ``D-'' in the model name is to indicate that we adapt them to the domain imbalance model.

Among these approaches, D-DRW and D-DRS are re-sampling and re-weighting methods with a deferred training scheduler. As suggested by \citet{cao2019learning} the re-sampling or re-weighting are only effective after 80\% of epochs have been trained. D-focal is a regularization-based method that uses an carefully designed loss function tailored for imbalanced data. D-BBN and D-HybridSC are two recent parameter isolation approaches that have shown state-of-the-art performance. D-MDFEND is used for multi-domain fake news detection which applies mixture-of-experts to deal with multi-domain transfer and isolation.

Regarding the DCMI hyperparameters i.e. ($\lambda_1$, $\lambda_2$), we used $(50,6)$, $(30,15)$, and $(4,3)$ for the ASC, DSC, and RFD datasets, respectively. Refer to Section~\ref{sec:Hyperparameters} for the hyperparameter search space and other implementation details.


\subsection{Quantitative Results}

\begin{table*}[t]
\centering
\resizebox{0.99\textwidth}{!}{
\begin{tabular}{ccc|cc|cc|cc}
\specialrule{.2em}{.1em}{.1em}
\multirow{2}{*}{Model} & \multicolumn{2}{c}{DSC} & \multicolumn{2}{c}{ASC} & \multicolumn{2}{c|}{RFD} & \multicolumn{2}{c}{Altered ASC} \\
 & Macro & Micro & Macro & Micro & Macro & Micro & Macro & Micro  \\
 \specialrule{.1em}{.05em}{.05em}
MTL (multitask learning) & 74.1{\footnotesize$\pm${3.1}} & 77.3{\footnotesize$\pm${3.8}} & 80.0{\footnotesize$\pm${1.8}} & 84.1{\footnotesize$\pm${0.7}} & 57.4* & 59.1* & 76.3{\footnotesize$\pm${2.9}} & 84.9{\footnotesize$\pm${2.4}}\\
D-AL (domain agnostic) & 80.6{\footnotesize$\pm${3.0}} & 81.3{\footnotesize$\pm${3.0}} & 82.5{\footnotesize$\pm${2.3}} & 84.8{\footnotesize$\pm${1.7}} & 68.8{\footnotesize$\pm${2.9}} & 70.2{\footnotesize$\pm${2.6}} & 51.9{\footnotesize$\pm${1.0}} & 61.1*  \\
D-DRS \citep{cao2019learning} & 76.3* & 76.6* & 84.3{\footnotesize$\pm${2.7}} & 86.0{\footnotesize$\pm${2.3}} & 71.4{\footnotesize$\pm${1.2}} & 72.6{\footnotesize$\pm${0.9}} & 51.4{\footnotesize$\pm${0.9}} & 58.3* \\
D-DRW \citep{cao2019learning} & 80.6{\footnotesize$\pm${3.4}} & 80.9{\footnotesize$\pm${3.2}} & 76.7* & 78.0* & 72.6{\footnotesize$\pm${0.8}} & 74.0{\footnotesize$\pm${0.6}} & 51.6{\footnotesize$\pm${1.2}} & 59.1*   \\
D-Focal \citep{lin2017focal} & 74.84* & 74.97* & 75.2* & 77.1* & 71.4{\footnotesize$\pm${3.2}} & 72.0{\footnotesize$\pm${3.4}} & 50.8{\footnotesize$\pm${0.5}} & 56.7*  \\
D-BBN \citep{zhou2020bbn} & 79.2{\footnotesize$\pm${3.7}} & 79.8{\footnotesize$\pm${3.8}} & 75.6* & 77.6* & 64.3* & 66.1* & 49.9{\footnotesize$\pm${1.4}} & 54.5{\footnotesize$\pm${3.9}} \\
D-HybridSC \citep{wang2021contrastive} & 82.4* & 82.4{\footnotesize$\pm${3.9}} & 83.5{\footnotesize$\pm${2.2}} & 84.9{\footnotesize$\pm${2.2}}  & 71.2{\footnotesize$\pm${1.4}} & 72.3{\footnotesize$\pm${1.2}} & 50.7{\footnotesize$\pm${1.0}} & 56.7* \\
D-MDFEND \citep{nan2021mdfend} & 80.5{\footnotesize$\pm${3.5}} & 80.8* & 81.0{\footnotesize$\pm${3.6}} & 82.8{\footnotesize$\pm${3.4}}  & 69.5{\footnotesize$\pm${2.0}} & 72.0{\footnotesize$\pm${2.5}} & 73.8* & 83.4* \\
 \hline
\textbf{DCMI (this work)} & \textbf{83.7}{\footnotesize$\pm${1.3}} & \textbf{83.8}{\footnotesize$\pm${1.3}} & \textbf{85.0}{\footnotesize$\pm${0.7}} & \textbf{87.2}{\footnotesize$\pm${0.4}} & \textbf{74.2}{\footnotesize$\pm${1.2}} & \textbf{74.1}{\footnotesize$\pm${1.0}} & \textbf{77.8}{\footnotesize$\pm${1.9}} & \textbf{85.2}{\footnotesize$\pm${1.4}} \\
\specialrule{.1em}{.05em}{.05em}
\multicolumn{9}{l}{\small * indicates that we only report the average results and there is a convergence issue due to the small training set or extreme imbalance} \\

\end{tabular}
}
\caption{Comparison of macro and micro averaged AUC results for DCMI (this work) and other baselines.
} 

\label{tab:overall}
\end{table*}

\subsubsection{Comparison with Other Work}
Table~\ref{tab:overall} presents a comparison of DCMI with other baselines. From this table, DCMI consistently outperforms other competitors for both metrics. 
Specifically, DCMI is much more data-efficient compared to other baselines, as it effectively encourages positive knowledge transfer across domains.
Among the three datasets, DCMI has the largest improvement margin for RFD. This can be attributed to the fact that domains in RFD are more diverse than those in ASC and DSC. The sentiment classification domains as in ASC and DSC have similarities as in these tasks positive or negative sentiments are usually expressed with similar words/phrases. For example, wonderful and terrible have similar interpretation for different tasks/domains to express positive or negative sentiment. 
However, expressions in fake news or rumors are far more diversified, follow more complex semantics, and even contradicting at times. For example, ``guns'' and ``shooting'' appear many times in ``Charlie Hebdo'' domain while they almost never appear in other domains like ``Germanwings Flight''. Even more interestingly, ``Trump'' appears frequently in both the fake news of ``COVID-19'' domains and the real news of ``government'', therefore it is a significant keyword with different domain interpretation. Under such domain disparities, selectively transferring common knowledge while preventing negative transfer becomes crucial which we believe is addressed by this work.


For the most recent state-of-the-art methods presented in Table~\ref{tab:overall}, we can observe mixed MIL performance results for different datasets indicating less adaptability compared to DCMI. This is perhaps because they do not employ any viable mechanism to explicitly encourage positive transfer. 

\subsubsection{Extremely Dissimilar Data} 
\label{sec:asc_altered}
We claim that DCMI is capable of adaptively selecting the useful knowledge (neurons) for a given domain and thus robust to extremely dissimilar domains. To demonstrate this, we create an artificial case where domains are extremely dissimilar in the dataset by design. Specifically, we divide the ASC dataset into two parts. The first part contains first 10 domains and the second part contains the other 9 domains. We keep the first part as is, while inverting the labels for the second part (i.e., flipping positive to negative and vice versa). Note that in a sentiment classification task such as ASC, domains are highly correlated so inverting labels for half of domains creates a drastic domain disparity.

Table \ref{tab:overall} shows the results of using the altered ASC data. We can see all baselines except MTL and D-MDFEND reach on only around 50\% AUC. This is because the extremely high domain divergence is causing a severe negative transfer and making it difficult for the majority of baselines to learn a good predictor. However, MTL and D-MDFEND perform better than other baselines, perhaps since negative transfer is reduced due to the use of separate heads for different domains in MTL and mixture-of-experts in D-MDFEND. Nevertheless, DCMI still outperforms MTL and D-MDFEND, confirming that DCMI is not only capable of isolating domain-specific knowledge but also is able to encourage positive transfer among similar domains, which is here for domains within each data part of the altered dataset.

\subsubsection{Ablation Study} 

\begin{table}[t]
\centering
\resizebox{\columnwidth}{!}{
\begin{tabular}{ccc|cc|cc}
\specialrule{.2em}{.1em}{.1em}
\multirow{2}{*}{Model} & \multicolumn{2}{c}{DSC} & \multicolumn{2}{c}{ASC} & \multicolumn{2}{c}{RFD} \\
 & Macro & Micro & Macro & Micro & Macro & Micro  \\
 \specialrule{.1em}{.05em}{.05em}
\textbf{DCMI} & \textbf{83.7}{\footnotesize$\pm${1.3}} & \textbf{83.8}{\footnotesize$\pm${1.3}} & \textbf{85.0}{\footnotesize$\pm${0.7}} & \textbf{87.2}{\footnotesize$\pm${0.4}} & \textbf{74.2}{\footnotesize$\pm${1.2}} & 74.1{\footnotesize$\pm${1.0}} \\
\hline
-$\mathcal{L}_{\text{dom}}$ & 81.9{\footnotesize$\pm${3.0}} & 82.3{\footnotesize$\pm${2.7}} & 84.5{\footnotesize$\pm${1.3}} & 86.7{\footnotesize$\pm${0.9}} & 73.1{\footnotesize$\pm${1.7}} & \textbf{74.3}{\footnotesize$\pm${0.8}} \\
-$\mathcal{L}_{\text{dom}},\mathcal{L}_{\text{con}}$ & 80.2{\footnotesize$\pm${3.4}} & 81.0{\footnotesize$\pm${3.2}} & 82.8{\footnotesize$\pm${1.6}} & 85.3{\footnotesize$\pm${1.4}} & 69.5{\footnotesize$\pm${1.3}} & 69.2{\footnotesize$\pm${0.9}} \\
\specialrule{.1em}{.05em}{.05em}
\end{tabular}
}
\caption{Ablation study of DCMI. "-$\mathcal{L}_{\text{dom}}$" and "$-\mathcal{L}_{\text{con}}$" indicate omitting the domain classification and contrastive loss terms, respectively.}

\label{tab:ablation}
\end{table}

\begin{table*}[h]
\centering
\resizebox{0.9\textwidth}{!}{
\begin{tabular}{ccc|cccc}
\specialrule{.2em}{.1em}{.1em}
Domains & Review & Label & D-AL &  \begin{tabular}[c]{@{}c@{}}DCMI\\-$\mathcal{L}_{\text{dom}},\mathcal{L}_{\text{con}}$\end{tabular}  & DCMI \\
 \specialrule{.1em}{.05em}{.05em}
Laptop & The nicest  part is the low heat output  and ultra quiet \textit{operation}. & P. & N. & P. & P.   \\ 
MicroMP3 & The flaw is inside the \textit{Zen}. & N. & P. & N. & N.   \\
Laptop & It feels cheap, the \textit{keyboard} is not very sensitive. & N. & P. & P. & N. \\ 
Restaurant & The downstairs bar scene is very cool and  chill...& P. & N. & N. & P.  \\ 
Restaurant & The \textit{sushi} is cut in blocks bigger than my   cell phone. & N. & P. & P. & N.   \\ 
\specialrule{.1em}{.05em}{.05em}
\end{tabular}
}
\caption{Qualitative comparison of predictions for different methods on a set of selected test samples from the ASC dataset \citep{DBLP:conf/naacl/KeXL21}. \textit{Italic} text indicates the aspect in the review. ``P.'' indicates positive and ``N.'' indicates negative assignments.
}

\label{tab:case}
\end{table*}

We conduct an ablation study to analyze the impact of each objective term.
The results of this experiment are presented in Table \ref{tab:ablation}. Here, ``-$\mathcal{L}_{\text{dom}}$'' indicates DCMI without the domain classification. ``-$\mathcal{L}_{\text{dom}},\mathcal{L}_{\text{con}}$'' indicates DCMI without the domain classification and contrastive loss. 
Note that if we remove the domain-aware representation layer in addition to $\mathcal{L}_{\text{dom}}$ and $\mathcal{L}_{\text{con}}$, DCMI becomes D-AL.
Based on the results provided in Table \ref{tab:ablation} the full DCMI system gives the best results, showing that every suggested component is crucial to the final model performance.

\subsection{Qualitative Results}

Table \ref{tab:case} shows several examples from ASC test set. For each example, we show the ground truth label (the third column), predictions of D-AL, DCMI and DCMI-[$\mathcal{L}_{\text{dom}},\mathcal{L}_{\text{con}}$]. By comparing D-AL and DCMI-[$\mathcal{L}_{\text{dom}},\mathcal{L}_{\text{con}}$], we can see the effectiveness of the domain-aware representation layer. By comparing DCMI and DCMI-[$\mathcal{L}_{\text{dom}},\mathcal{L}_{\text{con}}$], we can see whether the contrastive knowledge transfer is successful.

In the first row, ``quiet'' is a positive sentiment word in the ``laptop'' domain. However, ``quite'' can indicate negative in other domains (e.g., a ``quite'' earbud in ``MP3'' domain indicates negative sentiment). We can see DCMI and DCMI -[$\mathcal{L}_{\text{dom}},\mathcal{L}_{\text{con}}$] are able to separate the different polarity of the same sentiment word from different domains, while D-AL fails, suggesting that the knowledge selection in DCMI is capable of learning discriminative domain-aware representation. 

In the second row, we can see D-AL mistakenly takes the review as positive due to the small amount of training data in the ``MP3'' domain. DCMI and DCMI -[$\mathcal{L}_{\text{dom}},\mathcal{L}_{\text{con}}$] can make the correct prediction because of their ability to transfer knowledge from similar domains.

The last three rows of Table \ref{tab:case} showcase where only DCMI is correct. 
In the ``laptop'' domain (the third row), ``cheap'' conveys a negative sentiment in the example. However, ``cheap'' can indicate positive sentiment in the ``laptop'' domain if it is talking about the software domain. Therefore, an MIL model that only considers the annotated domain (e.g., DCMI-[$\mathcal{L}_{\text{dom}},\mathcal{L}_{\text{con}}$]) fails.
Similarly, the polarities of ``cool'' and ``chill'' depend not only on the dataset provided domain but also on the degrees of domain relevance for a given sample. 
The last case is an ironic expression, indicating DCMI provides a deeper understanding of the review.

In addition to the presented results, we provide a visual analysis of the domain-aware representation layer using t-SNE in the appendix.

\section{Conclusion}
In this work, we studied the problem of learning from multi-domain imbalanced data, where not only there is class imbalance but also there is an imbalance among domains with varying degrees of similarity. 
We proposed a novel technique called DCMI that is capable of identifying the \textit{shared knowledge} that can be transferred to improve the tail domain performance and the \textit{domain-specific knowledge} that needs to be handled carefully to avoid negative transfer. 
DCMI employs a domain-aware representation layer to adaptively select the relevant knowledge for each domain and leverages a novel contrastive learning objective to encourage knowledge transfer for relevant domains. 
Based on the experiments using three challenging multi-domain imbalanced datasets, DCMI shows improvements over the current state-of-the-art and demonstrates applicability to different scenarios.

\bibliography{refs}
\bibliographystyle{./template/acl_natbib}

\clearpage
\appendix

\section{Detailed Datasets Statistics}
\label{sec:dataset_detail}

In Table \ref{tab:dataset_asc}, \ref{tab:dataset_dsc}, and \ref{tab:dataset_rfd}, we
provide the frequency of samples corresponding to each domain for the ASC, DSC, and RFD datasets.


\begin{table}[h]
\centering
\resizebox{0.9\columnwidth}{!}{
\begin{tabular}{ccccccc}
\specialrule{.2em}{.1em}{.1em}
\multirow{2}{*}{Domains} & \multicolumn{2}{c}{Train} & \multicolumn{2}{c}{Validation} & \multicolumn{2}{c}{Test} \\
 & N. & P. & N. & P. & N. & P. \\
\specialrule{.1em}{.05em}{.05em}
Luxury Beauty & 1 & 2 & 1 & 1 & 260 & 2780 \\
Electronics & 61 & 436 & 7 & 54 & 773 & 5459 \\
CDs Vinyl & 7 & 99 & 1 & 12 & 89 & 1243 \\
Appliances & 1 & 1 & 1 & 1 & 3 & 184 \\
Digital Music & 1 & 12 & 1 & 1 & 401 & 15883 \\
AMAZON FASHION & 1 & 1 & 1 & 1 & 21 & 262 \\
Office Products & 4 & 55 & 1 & 6 & 55 & 693 \\
Books & 146 & 1835 & 18 & 229 & 1834 & 22946 \\
Gift Cards & 1 & 1 & 1 & 1 & 4 & 290 \\
Grocery Gourmet Food & 7 & 77 & 1 & 9 & 91 & 970 \\
Cell Phones Accessories & 11 & 71 & 1 & 8 & 138 & 890 \\
Prime Pantry & 1 & 9 & 1 & 1 & 692 & 12160 \\
Home Kitchen & 53 & 457 & 6 & 57 & 663 & 5719 \\
Magazine Subscriptions & 1 & 1 & 1 & 1 & 22 & 192 \\
Pet Supplies& 19 & 133 & 2 & 16 & 244 & 1673 \\
Software & 1 & 1 & 1 & 1 & 222 & 899 \\
Sports Outdoors & 17 & 193 & 2 & 24 & 212 & 2415 \\
All Beauty & 1 & 1 & 1 & 1 & 18 & 498 \\
Automotive  & 10 & 118 & 1 & 14 & 132 & 1475 \\
Musical Instruments & 1 & 16 & 1 & 2 & 1475 & 20058 \\
Movies TV & 29 & 215 & 3 & 26 & 365 & 2693 \\
Video Games & 4 & 31 & 1 & 3 & 5502 & 39327 \\
Tools Home Improvement & 13 & 140 & 1 & 17 & 170 & 1760 \\
Toys Games & 9 & 126 & 1 & 15 & 121 & 1575 \\
Patio Lawn Garden & 6 & 52 & 1 & 6 & 83 & 656 \\
Arts Crafts Sewing & 2 & 35 & 1 & 4 & 2714 & 43844 \\
Clothing Shoes Jewelry & 89 & 726 & 11 & 90 & 1120 & 9075 \\
Kindle Store & 9 & 152 & 1 & 19 & 115 & 1909 \\
Industrial Scientific  & 1 & 5 & 1 & 1 & 442 & 6821 \\
 \specialrule{.1em}{.05em}{.05em}
\end{tabular}
}
\caption{The number of samples in each domain and data split for the DSC dataset. ``N.'' indicates negative labels and ``P.'' indicates positive labels.}
\label{tab:dataset_asc}
\vspace{-3mm}
\end{table}

\begin{table}[h]
\centering
\resizebox{\columnwidth}{!}{
\begin{tabular}{cccccccc}
\specialrule{.2em}{.1em}{.1em}
\multirow{2}{*}{Dataset} & \multirow{2}{*}{Domains} & \multicolumn{2}{c}{Train} & \multicolumn{2}{c}{Validation} & \multicolumn{2}{c}{Test} \\
 &  & \begin{tabular}[c]{@{}c@{}}Fake/\\ Rumour\end{tabular} & Real & \begin{tabular}[c]{@{}c@{}}Fake\\ /Rumour\end{tabular} & Real & \begin{tabular}[c]{@{}c@{}}Fake/\\ Rumour\end{tabular} & Real \\
\specialrule{.1em}{.05em}{.05em}

\multirow{5}{*}{PHEME} & Ferguson  & 51 & 17 & 8 & 2 & 258 & 86 \\
 & Charlie Hebdo & 97 & 27 & 16 & 4 & 487 & 138 \\
 & \begin{tabular}[c]{@{}c@{}}Germanwings \\ Crash\end{tabular} & 13 & 14 & 2 & 2 & 70 & 72 \\
 & Sydney Siege & 41 & 31 & 7 & 5 & 210 & 157 \\
 & Ottawa Shooting & 25 & 28 & 4 & 4 & 126 & 141 \\
\specialrule{.1em}{.05em}{.05em}
LIAR & Politic  & 199 & 168 & 26 & 16 & 250 & 211 \\
\specialrule{.1em}{.05em}{.05em}
\end{tabular}
}
\caption{The number of samples in each domain and data split for the RFD dataset. RFD is composed of PHEME and LIAR data. ``N.'' indicates negative labels and ``P.'' indicates positive labels.}
\label{tab:dataset_rfd}
\vspace{-3mm}
\end{table}

\begin{table}[]
\centering
\resizebox{\columnwidth}{!}{
\begin{tabular}{cccccccc}
\specialrule{.2em}{.1em}{.1em}
\multirow{2}{*}{Dataset} & \multirow{2}{*}{Domains} & \multicolumn{2}{c}{Train} & \multicolumn{2}{c}{Validation} & \multicolumn{2}{c}{Test} \\
 &  & N. & P. & N. & P. & N. & P. \\
\specialrule{.1em}{.05em}{.05em}
\multirow{2}{*}{SemEval14} & laptop & 80 & 93 & 66 & 57 & 128 & 341 \\
 & restaurant & 77 & 209 & 26 & 70 & 196 & 728 \\
\specialrule{.1em}{.05em}{.05em}
\multirow{9}{*}{Ding9Domains} & HitachiRouter & 3 & 9 & 9 & 18 & 32 & 74 \\
 & CanonS100 & 4 & 6 & 2 & 20 & 11 & 77 \\
 & ipod & 3 & 6 & 6 & 13 & 20 & 57 \\
 & Nokia6600 & 8 & 14 & 15 & 30 & 48 & 134 \\
 & DiaperChamp & 2 & 9 & 5 & 19 & 26 & 70 \\
 & CanonD500 & 1 & 5 & 1 & 13 & 8 & 52 \\
 & Norton & 2 & 9 & 8 & 16 & 38 & 60 \\
 & MicroMP3 & 21 & 9 & 45 & 15 & 170 & 73 \\
 & LinksysRouter & 7 & 3 & 20 & 2 & 59 & 30 \\
\specialrule{.1em}{.05em}{.05em}
\multirow{5}{*}{HL5Domains} & Creative40G & 14 & 28 & 35 & 50 & 155 & 184 \\
 & ApexAD2600 & 12 & 9 & 26 & 17 & 87 & 85 \\
 & Nokia6610 & 11 & 5 & 28 & 6 & 114 & 22 \\
 & Nikon4300 & 8 & 1 & 16 & 4 & 74 & 8 \\
 & CanonG3 & 11 & 2 & 21 & 7 & 89 & 26 \\
\specialrule{.1em}{.05em}{.05em}
\multirow{3}{*}{Liu3Domains} & Computer & 13 & 4 & 34 & 1 & 101 & 41 \\
 & Router & 9 & 6 & 19 & 12 & 73 & 50 \\
 & Speaker & 19 & 2 & 31 & 13 & 140 & 36 \\
 \specialrule{.1em}{.05em}{.05em}
\end{tabular}
}
\caption{The number of samples in each domain and data split for the ASC dataset. ASC is composed of four datasets. ``N.'' indicates negative labels and ``P.'' indicates positive labels.}
\label{tab:dataset_dsc}
\vspace{-3mm}
\end{table}

\clearpage
\section{Visual Analysis of the Domain-aware Representation Layer} 
\begin{figure}[h]
     \centering
     \begin{subfigure}[b]{\columnwidth}
         \centering
         \includegraphics[width=0.9\textwidth]{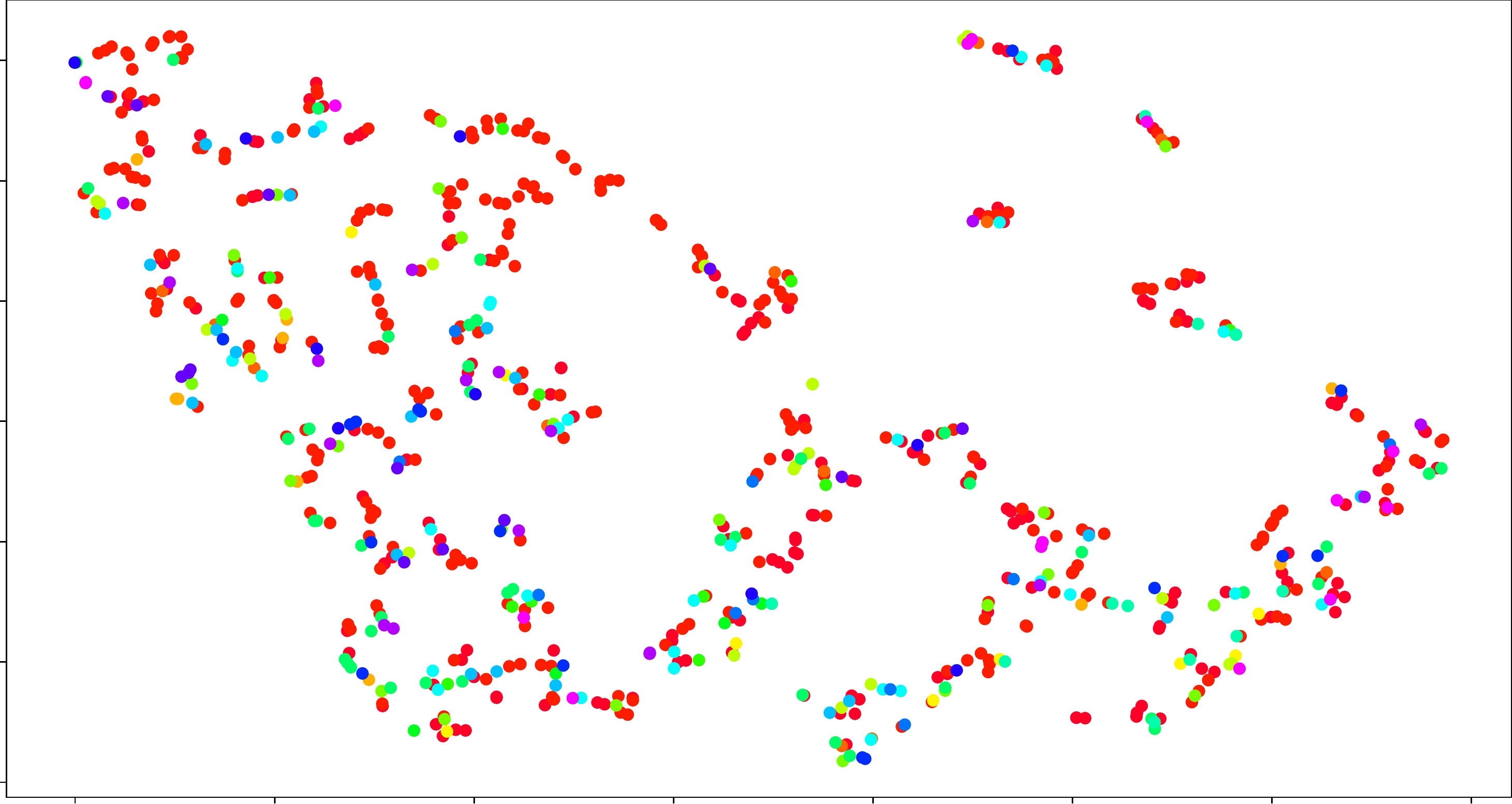}
         \caption{Before domain-aware representation layer}
         \label{fig:before}
     \end{subfigure}
     \hfill
     \begin{subfigure}[b]{\columnwidth}
         \includegraphics[width=\textwidth]{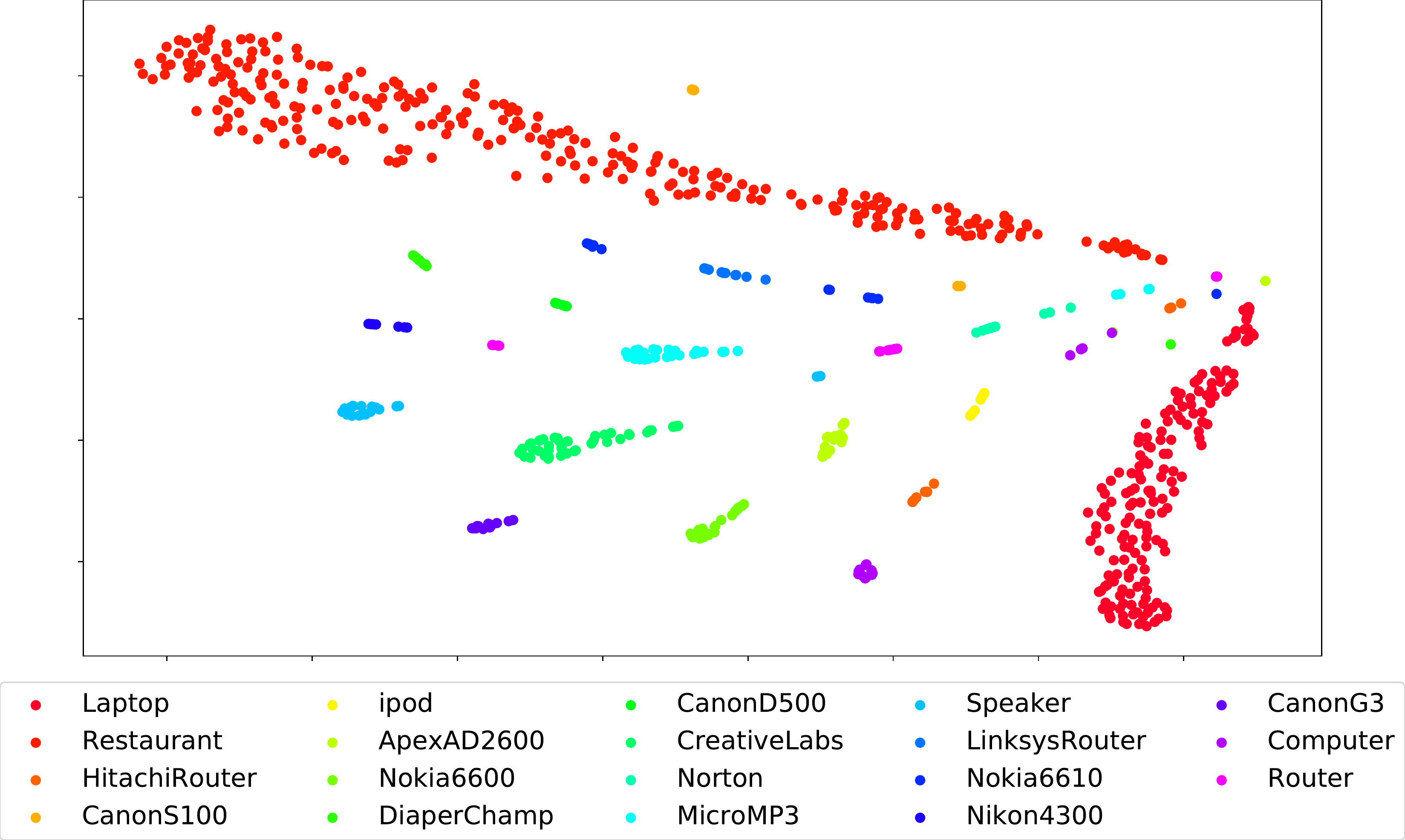}
         \caption{After domain-aware representation layer}
         \label{fig:after}
     \end{subfigure}
    \caption{t-SNE visualization of sample representation for different domains, (a) before and (b) after the domain-aware representation layer for the ASC dataset \citep{DBLP:conf/naacl/KeXL21}. Figure best viewed in color.}
    \label{fig:tsne}     
\end{figure}

We visualize sample representations before and after the domain-aware representation layer using for ASC dataset. See Figure \ref{fig:tsne} for t-SNE visualizations.
Here, we color the samples according to their domain assignments. 

Before the domain-aware representation layer, we can see the points related to different domains are mixed and hard to differentiate. However, after the domain-aware representation layer, samples with similar colors form clusters, indicating a higher embedding distance for different domains. From this visualization, we can infer that the suggested method is able to learn discriminative domain-aware representations.

\end{document}